\documentclass[nohyperref]{article}

\usepackage[margin=1in]{geometry}
\usepackage{mathpazo}

\usepackage{amsmath,amssymb,amsthm,amsfonts,bm}
\usepackage{xspace}
\usepackage{graphicx}
\usepackage{color}
\usepackage{booktabs}
\usepackage[colorlinks=true,linkcolor=blue,citecolor=blue,urlcolor=blue]{hyperref}

\usepackage{todonotes}
\usepackage{subcaption}
\usepackage{url}
\usepackage{comment}
\usepackage{algorithm,algorithmic}
\usepackage[toc,page,header]{appendix}
\usepackage{wrapfig}

\usepackage[T1]{fontenc}
\usepackage[backend=biber,style=alphabetic,natbib=true,maxbibnames=99,minalphanames=3]{biblatex}
\newcommand{\xd}{\mathbf{x}}
\newcommand{\zd}{\mathbf{z}}
\newcommand{\ed}{\bm{\varepsilon}}
\usepackage[normalem]{ulem}
\usepackage{xcolor}

\newcommand{\scrcmd}{\textsuperscript}
\newcommand{\scr}[1]{\scrcmd{#1}}

\newcommand{\dalle}{DALL$\cdot$E~2}

\title{Raising the Cost of Malicious AI-Powered Image Editing}
\author{
    Hadi Salman\footnote{Equal contribution.} \\
    \texttt{hady@mit.edu} \\
    MIT
    \and
    Alaa Khaddaj\footnotemark[1] \\
    \texttt{alaakh@mit.edu} \\
    MIT
    \and
    Guillaume Leclerc\footnotemark[1] \\
    \texttt{gleclerc@mit.edu} \\
    MIT
    \and
    Andrew Ilyas \\
    \texttt{ailyas@mit.edu} \\
    MIT
    \and
    Aleksander M\k{a}dry \\
    \texttt{madry@mit.edu} \\
    MIT
}
\date{}

\begin{document}
\makeatletter 
\let\c@table\c@figure
\let\c@lstlisting\c@figure
\makeatother

    \maketitle

    \begin{abstract}
        We present an approach to mitigating the risks of malicious image editing posed by large diffusion models. The key idea is to \textit{immunize} images so as to make them resistant to manipulation by these models. This immunization relies on injection of imperceptible adversarial perturbations designed to disrupt the operation of the targeted diffusion models, forcing them to generate unrealistic images. We provide two methods for crafting such perturbations, and then demonstrate their efficacy. Finally, we discuss a policy component necessary to make our approach fully effective and practical---one that involves the organizations developing diffusion models, rather than individual users, to implement (and support) the immunization process.\footnote{Code is available at \url{https://github.com/MadryLab/photoguard}.}

    \end{abstract}

    \section{Introduction}
    \label{sec:intro}
\begin{figure*}[!t]
    \centering
    \includegraphics[width=\linewidth]{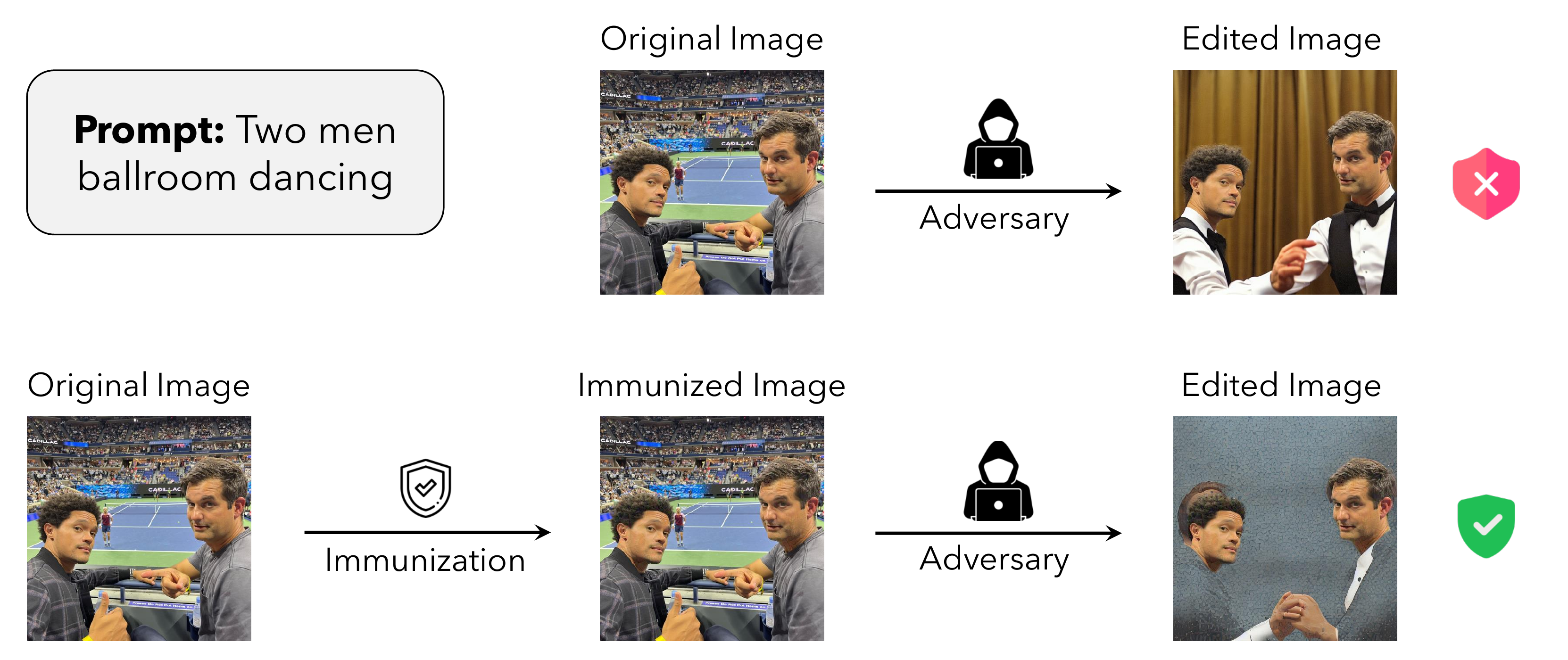}
    \caption{{\em Overview of our framework}. An adversary seeks to modify an image found online. The adversary describes via a textual prompt the desired changes and then uses a diffusion model to generate a realistic image that matches the prompt~{(top)}. By immunizing the original image before the adversary can access it, we disrupt their ability to successfully perform such edits (bottom).}
    \label{fig:framework}
\end{figure*}

    Large diffusion models such as \dalle{} \cite{ramesh2022hierarchical}
and Stable Diffusion~\cite{rombach2022high}
are known for their ability to produce high-quality photorealistic
images, and can be used for a variety of image synthesis and editing tasks.
However, the ease of use of these models has raised concerns about their
potential abuse, e.g., by creating inappropriate or harmful digital content.
For example, a malevolent actor might download photos of people posted online and edit them maliciously using an off-the-shelf diffusion model
(as in Figure \ref{fig:framework} top).

How can we address these concerns? First, it is important to recognize that it is, in some sense, impossible to completely eliminate such malicious image editing.
Indeed, even without diffusion models in the picture, malevolent actors can
still use tools such as Photoshop to manipulate existing images,
or even synthesize fake ones entirely from scratch.
The key new problem that large generative models introduce is
that these actors can
now create realistic edited images with {\em ease}, i.e., without the need for specialized skills or
expensive equipment. This realization motivates us to ask:
\begin{center}
    \textit{How can we raise the cost of malicious (AI-powered) image manipulation?}
\end{center}
In this paper, we put forth an approach that aims to alter the economics of
AI-powered image editing. At the core of our approach is the idea of image \textit{immunization}---that is, making a specific image resistant to AI-powered manipulation by adding a
carefully crafted (imperceptible) perturbation to it.
This perturbation would disrupt the operation of a diffusion model, forcing the edits it performs to be unrealistic (see Figure ~\ref{fig:framework}).
In this paradigm, people can thus continue to share their (immunized) images as usual, while getting a layer of protection against undesirable manipulation.

We demonstrate how one can craft such imperceptible perturbations
for large-scale diffusion models and show that they can indeed prevent realistic image editing. We then discuss in Section~\ref{sec:discussion} complementary technical and policy components needed to make our approach fully effective and practical.

    \section{Preliminaries}
    \label{sec:prelim}
    We start by providing an overview of diffusion models as well as of the key concept we will leverage: adversarial attacks.
\subsection{Diffusion Models}
\label{sec:diffusion_models_prelim}

Diffusion models have emerged recently as powerful tools for generating realistic images~\citep{sohl-dickstein15diffusion, ho2020ddpm}. These models excel especially at generating and editing images using textual prompts, and currently surpass other image generative models such as GANs~\citep{goodfellow2014generative} in terms of the quality of produced images.

\paragraph{Diffusion process.}
At their core, diffusion models employ a stochastic differential process called the {\em diffusion process}~\citep{sohl-dickstein15diffusion}. This process allows us to view the task of (approximate) sampling from a distribution of real images $q(\cdot)$ as a series of {\it denoising} problems. More precisely, given a sample $\xd_0 \sim q(\cdot)$, the diffusion process incrementally adds noise to generate samples $\xd_1, \ldots, \xd_T$ for $T$ steps, where $\xd_{t+1} = a_{t} \xd_{t} + b_{t} \ed_{t}$, and $\ed_{t}$ is sampled from a Gaussian distribution\footnote{Here, $a_{t}$ and $b_{t}$ are the parameters of the distribution $q(\xd_{t+1} | \xd_{t})$. Details are provided in Appendix~\ref{app:extended_background}.}. Note that, as a result, the sample $\xd_T$ starts to follow a standard normal distribution $\mathcal{N}(0,\textbf{I})$ when ${T\to\infty}$. Now, if we reverse this process and are able to sample $\xd_{t}$ given $\xd_{t+1}$, i.e., {\em denoise $\xd_{t+1}$}, we can ultimately generate new samples from $q(\cdot)$. This is done by simply starting from $\xd_T \sim \mathcal{N}(0, \textbf{I})$ (which corresponds to $T$ being sufficiently large), and iteratively denoising these samples for T steps, to produce a new image $\tilde{\xd} \sim q(\cdot)$.

The element we need to implement this process is thus to learn a neural network $\ed_\theta$ that ``predicts'' given $\xd_{t+1}$ the noise $\ed_t$ added to $\xd_t$ at each time step $t$. Consequently, this {\em denoising model} $\ed_\theta$ is trained to minimize the following loss function:

\begin{equation}
    \label{eq:diff_opt}
    \begin{aligned}
        \mathcal{L}(\theta) = \mathbb{E}_{t, \xd_0, \ed \sim \mathcal{N}(0,1)} \left[ \lVert \ed - \ed_{\theta}(\xd_{t+1}, t) \rVert_2^2 \right],
    \end{aligned}
\end{equation}

\noindent
where $t$ is sampled uniformly over the $T$ time steps. We defer discussion of details to Appendix~\ref{app:extended_background} and refer the reader to \cite{weng2021diffusion} for a more in-depth treatment of diffusion models.

\begin{figure*}[!t]
    \centering
    \includegraphics[width=.9\linewidth]{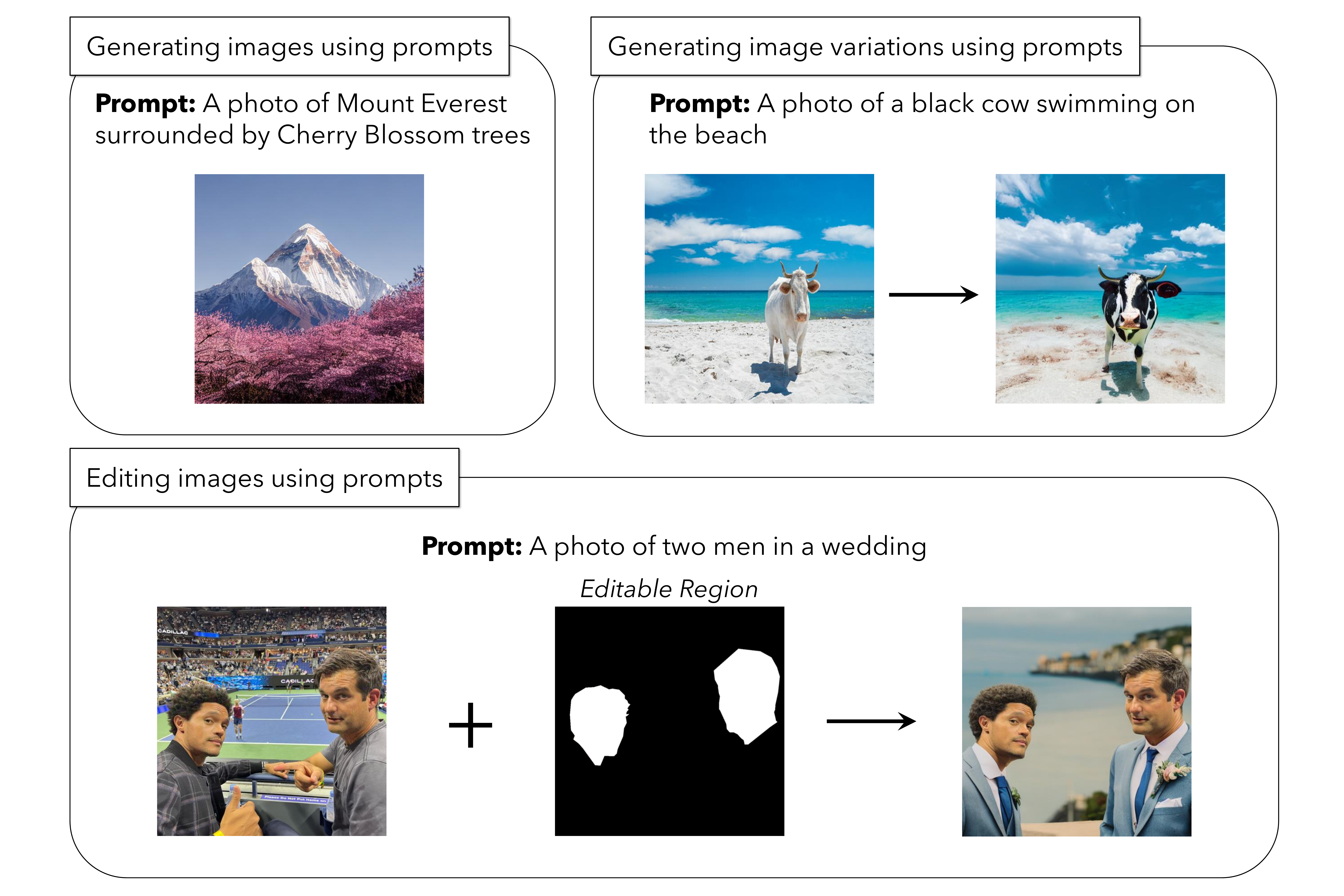}
    \caption{Diffusion models offer various capabilities, such as  (1) generating images using textual prompts (top left), (2) generating variations of an input image using textual prompts (top right), and (3) editing images using textual prompts (bottom).}
    \label{fig:applications_of_diff_models}
\end{figure*}
\paragraph{Latent diffusion models (LDMs).} 
Our focus will be on a specific class of diffusion models called the {\em latent diffusion models} (LDMs)~\citep{rombach2022high}\footnote{Our methodology can be adjusted to other diffusion models. Our focus on LDMs is motivated by the fact that all popular open-sourced diffusion models are of this type.}. These models apply the diffusion process
described above in the \textit{latent space} instead of the input (image) space. As it turned out, this change enables more efficient training and faster inference, while maintaining high quality generated samples.

Training an LDM is similar to training a standard diffusion model and differs mainly in one aspect. Specifically, to train an LDM, the input image $\xd_0$ is first mapped to its latent representation $\zd_0 = \mathcal{E}(\xd_0)$, where $\mathcal{E}$ is a given encoder. The diffusion process then continues as before (just in the {\em latent} space) by incrementally adding noise to generate samples $\zd_1, \ldots, \zd_T$ for $T$ steps, where $\zd_{t+1} = a_{t} \zd_{t} + b_{t} \ed_{t}$, and $\ed_{t}$ is sampled from a Gaussian distribution. Finally, the denoising network $\ed_{\theta}$ is then learned analogously to as before but, again, now in the latent space, by minimizing the following loss function:

\begin{equation}
    \label{eq:ldm_opt}
    \begin{aligned}
        \mathcal{L}(\theta) = \mathbb{E}_{t, \zd_0, \ed \sim \mathcal{N}(0,1)} \left[ \lVert \ed - \ed_{\theta}(\zd_{t+1}, t) \rVert_2^2 \right]
    \end{aligned}
\end{equation}

\noindent
Once the denoising network $\ed_{\theta}$ is trained, the same generative process can be applied as before, starting from a random vector in the latent space, to obtain a latent representation $\tilde{\zd}$ of the (new) generated image. This representation is then decoded into an image $\tilde{\xd} = \mathcal{D}(\tilde{\zd}) \sim q(\cdot)$, using the corresponding decoder~$\mathcal{D}$.

\paragraph{Prompt-guided sampling using an LDM.}
An LDM by default generates a random sample from the distribution of images  $q(\cdot)$ it was trained on. However, it turns out one can also guide the sampling using natural language. This can be accomplished by combining the latent representation $\zd_T$ produced during the diffusion process with the embedding of the user-defined textual prompt $t$.\footnote{Conditioning on the text embedding happens at every stage of the generation process. See \citep{rombach2022high} for more details.} The denoising network $\ed_\theta$ is applied to the combined representation for $T$ steps, yielding $\tilde{\zd}$ which is then mapped to a new image using the decoder $\mathcal{D}$ as before.

\paragraph{LDMs capabilities.} 
LDMs turn out to be powerful text-guided image generation and editing tools. In particular, LDMs can be used not only for generating images using textual prompts, as described above, but also for generating textual prompt--guided variations of an image or edits of a specific part of an image (see Figure~\ref{fig:applications_of_diff_models}). The latter two capabilities (i.e., generation of image variations and image editing) requires a slight modification of the generative process described above. Specifically, to modify or edit a given image $\xd$, we condition the generative process on this image. That is, instead of applying, as before, our generative process of $T$ denoising steps to a random vector in the latent space, we apply it to the latent representation obtained from running the latent diffusion process on our image $\xd$. To edit only part of the image we additionally condition the process on freezing the parts of the image that were to remain unedited.

\subsection{Adversarial Attacks}
\label{sec:adv_attacks_prelim}
For a given computer vision model and an image, an {\em adversarial example} is an
imperceptible perturbation of that image that manipulates the model's behavior~\citep{szegedy2014intriguing,biggio2013evasion}.
In image classification, for example, an adversary can
construct an adversarial example for a given image $\xd$ that makes it classified as a specific target label $y_{targ}$ (different from the true label). This construction is achieved by minimizing the loss of a classifier $f_\theta$ with respect to that image:
\begin{align}
    \label{eq:adversarial-example-opt}
    \delta_{adv} = \arg \min_{\delta \in \Delta} \mathcal{L}(f_\theta(\xd+\delta), y_{targ}).
\end{align}
Here, $\Delta$ is a set of perturbations that are small enough that they are imperceptible---a common choice is to
constrain the adversarial example to be close (in $\ell_p$ distance) to the original
image, i.e., $\Delta = \{ \delta: \lVert \delta \rVert_p \leq \epsilon\}$.
The canonical approach to constructing an adversarial example
is to solve the optimization problem~\eqref{eq:adversarial-example-opt} via projected gradient descent (PGD) \citep{nesterov2003introductory, madry2018towards}.

    \section{Adversarially Attacking Latent Diffusion Models}
    \label{sec:methodology}

\begin{figure*}[!t]
    \centering
    \includegraphics[width=\linewidth]{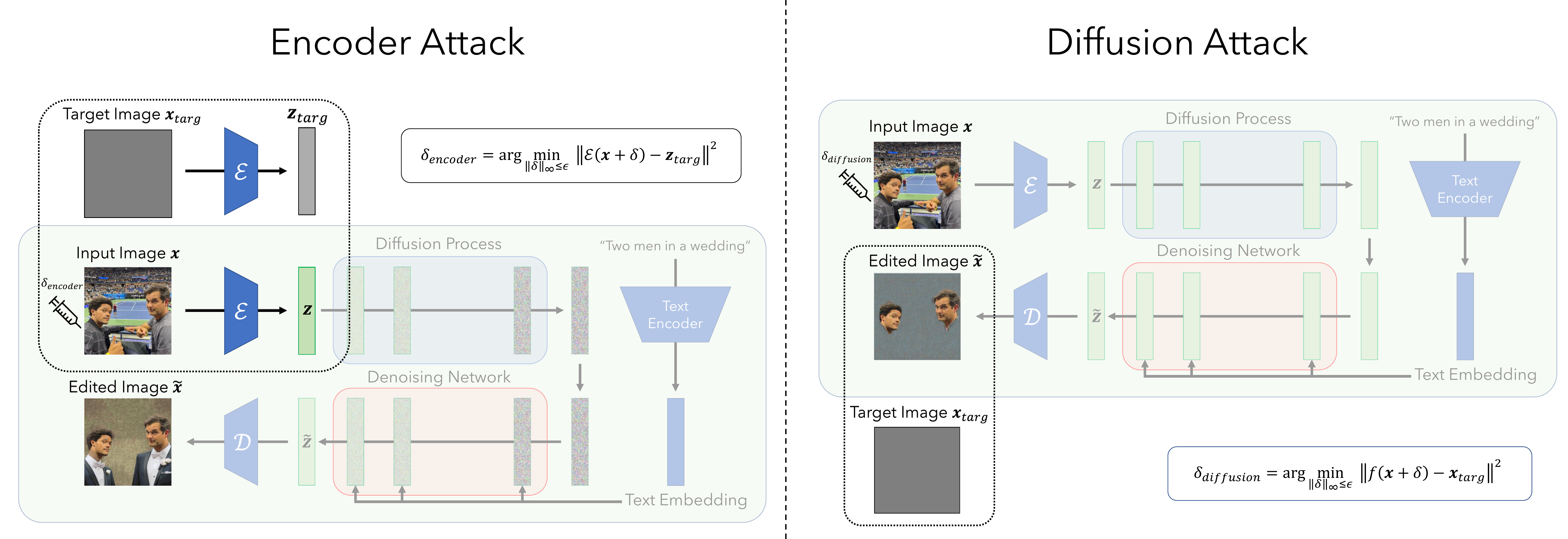}
    \caption{
        {\em Overview of our proposed attacks}. When applying the \textit{encoder} attack (left), our goal is to map the representation of the original image to the representation of a target image (gray image). Our (more complex) \textit{diffusion} attack (right), on the other hand, aims to break the diffusion process by manipulating the whole process to generate image that resembles a given target image (gray image).}
    \label{fig:attack_diag}
\end{figure*}

We now describe our approach to immunizing images, i.e., making them harder to manipulate using latent diffusion models (LDMs). At the core of our approach is to leverage techniques from the adversarial attacks literature~\citep{szegedy2014intriguing,madry2018towards,akhtar21advsurvey} and add adversarial perturbations (see Section~\ref{sec:diffusion_models_prelim}) to immunize images. Specifically, we present two different methods to execute this strategy (see Figure~\ref{fig:attack_diag}): an \textit{encoder attack}, and a \textit{diffusion attack}.

\paragraph{Encoder attack.}

Recall that an LDM, when applied to an image, first encodes the image using an encoder $\mathcal{E}$ into a latent vector representation, which is then used to generate a new image (see Section~\ref{sec:prelim}). The key idea behind our encoder attack is now to disrupt this process by forcing the encoder to map the input image to some ``bad'' representation. To achieve this, we solve the following optimization problem using projected gradient descent (PGD):
\begin{align}
    \delta_{encoder} = \arg \min_{\|\delta \|_\infty \leq \epsilon} \|\mathcal{E}(\xd+\delta) - \zd_{targ}\|_2^2,
\end{align}
where $\xd$ is the image to be immunized, and $\zd_{targ}$ is some target latent representation (e.g., $\zd_ {targ}$ can be the representation, produced using encoder $\mathcal{E}$, of a gray image). Solutions to this optimization problem yield small, imperceptible perturbations $\delta_{encoder}$ which, when added to the original image, result in an (immunized) image that is similar to the (gray) target image from the  LDM's encoder perspective. This, in turn, causes the LDM to generate an irrelevant or unrealistic image. An overview of this attack is shown in Figure~\ref{fig:attack_diag} (left)\footnote{See Algorithm~\ref{alg:simple_attack_alg} in Appendix for the details of the encoder attack.}.

\paragraph{Diffusion attack.} Although the encoder attack is effective at forcing the LDM to generate images that are unrelated to the immunized ones, we still expect the LDM to use the textual prompt. For example, as shown in the encoder attack diagram in Figure~\ref{fig:attack_diag}, editing an immunized image of two men using the prompt ``Two men in a wedding'' still results in a generated image with two men wearing wedding suits, even if the image will contain some visual artifacts indicating that it has been manipulated. Can we disturb the diffusion process even further so that the diffusion model ``ignores'' the textual prompt entirely and generates a more obviously manipulated image?

It turns out that we are able to do so by using a more complex attack, one where we target the diffusion process itself instead of just the encoder. In this attack, we perturb the input image so that the {\em final} image generated by the LDM is a specific target image (e.g., random noise or gray image). Specifically, we generate an adversarial perturbation $\delta_{diffusion}$ by solving the following optimization problem (again via PGD):
\begin{align}
    \label{eq:diffusion_attack}
    \delta_{diffusion} = \arg \min_{\|\delta \|_\infty \leq \epsilon} \|f(\xd+\delta) - \xd_{targ}\|_2^2.
\end{align}
Above,  $f$ is the LDM, $\xd$ is the image to be immunized, and $\xd_{targ}$ is the target image to be generated. An overview of this attack is depicted in Figure~\ref{fig:attack_diag} (right)\footnote{See Algorithm~\ref{alg:complex_attack_alg} in Appendix for the details  of the diffusion attack.}. As we already mentioned, this attack targets the full diffusion process (which includes the text prompt conditioning), and tries to nullify not only the effect of the immunized image, but also that of the text prompt itself. Indeed, in our example (see Figure~\ref{fig:attack_diag}~(right)) no wedding suits appear in the edited image whatsoever.

It is worth noting that this approach, although more powerful than the encoder attack, is harder to execute. Indeed, to solve the above problem~\eqref{eq:diffusion_attack} using PGD, one needs to backpropagate through the full diffusion process (which, as we recall from Section~\ref{sec:diffusion_models_prelim}, includes repeated application of the denoising step). This causes memory issues even on the largest GPU we used\footnote{We used an A100 with 40 GB memory.}. To address this challenge, we backpropagate through only a few steps of the diffusion process, instead of the full process, while achieving adversarial perturbations that are still effective. We defer details of our attacks to Appendix~\ref{app:experimental}.

    \section{Results}
    \label{sec:results}
    
In this section, we examine the effectiveness of our proposed immunization method.

\paragraph{Setup.} 
We focus on the Stable Diffusion Model (SDM) v1.5~\citep{rombach2022high}, though our methods can be applied to other diffusion models too. In each of the following experiments, we aim to disrupt the performance of SDM by adding imperceptible noise (using either of our proposed attacks)---i.e., applying our immunization procedure---to a variety of images. The goal is to force the model to generate images that are unrealistic and unrelated to the original (immunized) image. We evaluate the performance of our method both qualitatively (by visually inspecting the generated images) and quantitatively (by examining the image quality using standard metrics). We defer further experimental details to Appendix~\ref{app:experimental}.

\begin{figure*}[!t]
    \centering
    \includegraphics[width=.8\linewidth]{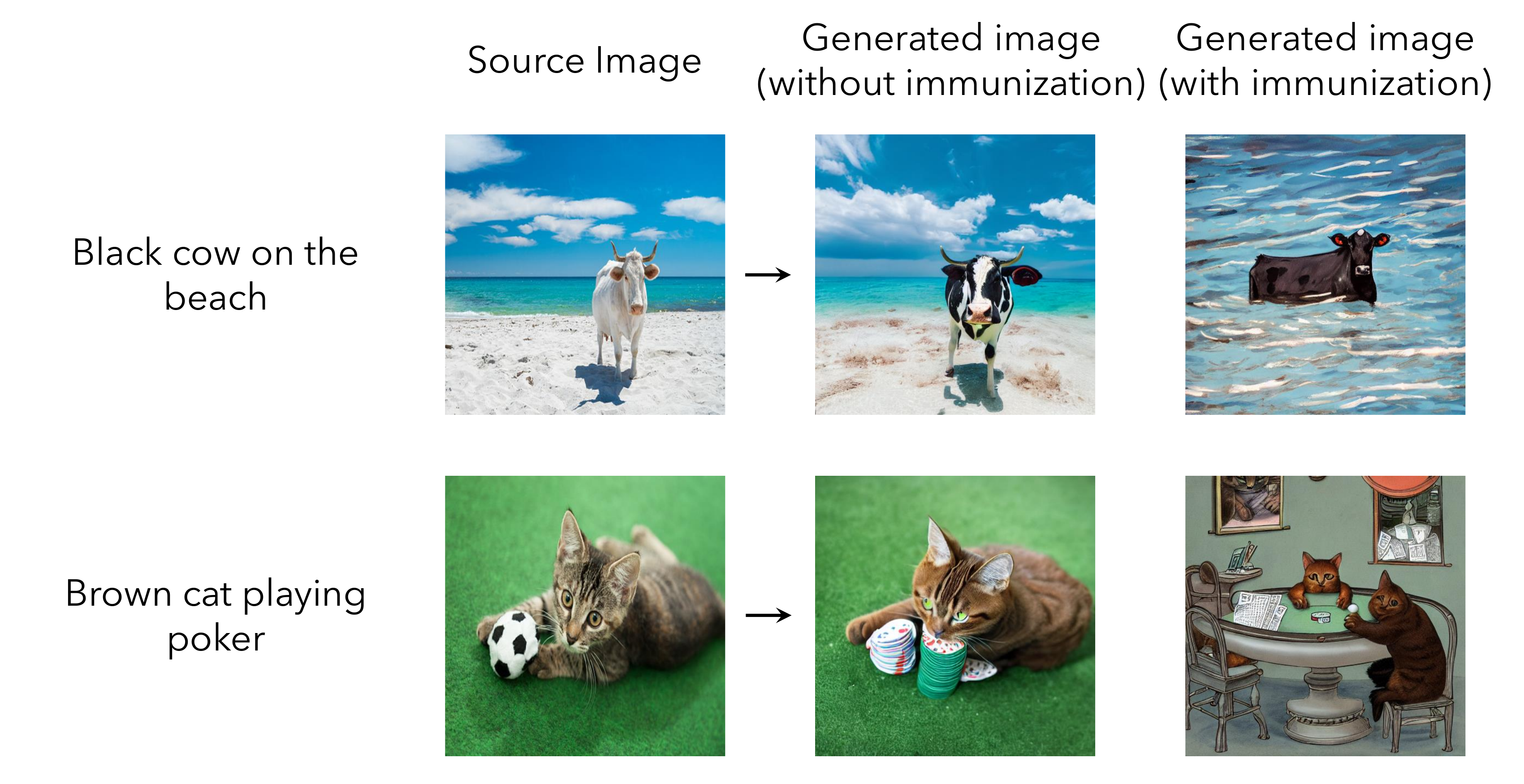}
    \caption{
    Given a source image (e.g., image of a white cow on the beach) and a textual prompt (e.g., "black cow on the beach"), the SDM can generate a realistic image matching the prompt while still similar to the original image (middle column). However, when the source image is immunized, the SDM fails to do so (right-most column). More examples are in Appendix~\ref{app:additional_results}.
    }
    \label{fig:image_var_simple_attack}
\end{figure*}

\subsection{Qualitative Results}
\paragraph{Immunizing against generating image variations.} We first assess whether we can disrupt the SDM's ability to generate realistic variations of an image based on a given textual prompt. For example, given an image of a white cow on the beach and a prompt of ``black cow on the beach'', the SDM should generate a realistic image of a \textit{black} cow on the beach that looks similar to the original one (cf. Figure~\ref{fig:image_var_simple_attack}). Indeed, the SDM is able to generate such images. However, when we immunize the original images (using the encoder attack), the SDM fails to generate a realistic variation---see Figure~\ref{fig:image_var_simple_attack}.

\begin{figure*}[!t]
    \centering
    \includegraphics[width=\linewidth]{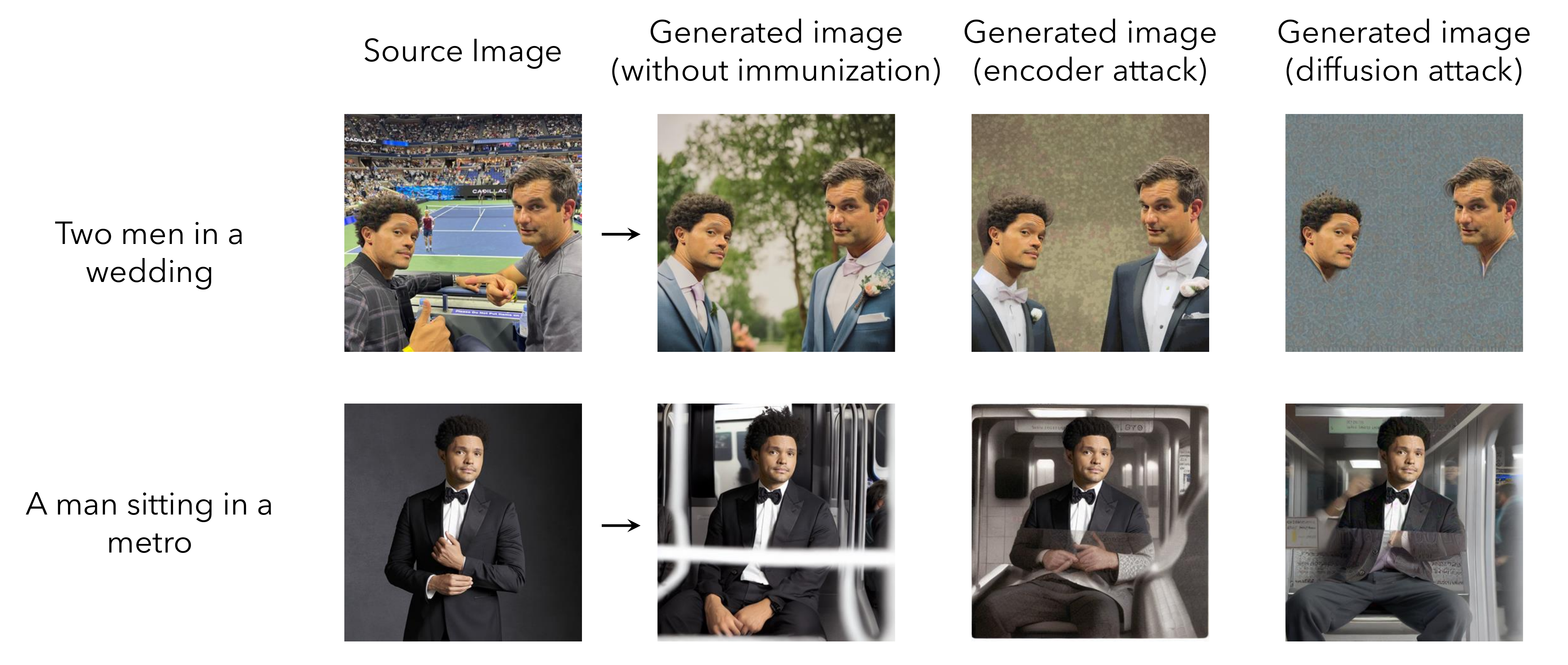}
    \caption{
        Given a source image (e.g., image of two men watching a tennis game) and a textual prompt (e.g., "two men in a wedding"), the SDM can edit the source image to match the prompt (second column). However, when the source image is immunized using the \textit{encoder attack}, the SDM fails to do so (third column). Immunizing using the \textit{diffusion attack} further reduces the quality of the edited image (forth column). More examples are in Appendix~\ref{app:additional_results}.}
    \label{fig:image_edit_attacks}
\end{figure*}

\paragraph{Immunizing against image editing.} 
Now we consider the more challenging task of disrupting the ability of SDMs to edit images using textual prompts. The process of editing an image using an SDM involves inputting the image, a mask indicating which parts of the image should be edited, and a text prompt guiding how the rest of the image should be manipulated. The SDM then generates an edited version based on that prompt. An example can be seen in Figure~\ref{fig:applications_of_diff_models}, where an image of two men watching a tennis game is transformed to resemble a wedding photo. This corresponded to inputting the original image, a binary mask excluding from editing only the men's heads, and the prompt ``A photo of two men in a wedding.'' However, when the image is immunized (using either encoder or diffusion attacks), the SDM is unable to produce realistic image edits (cf. Figure~\ref{fig:image_edit_attacks}). Furthermore, the diffusion attack results in more unrealistic images than the encoder attack.

\begin{table}[!t]
    \centering
    \resizebox{\columnwidth}{!}{%
    \begin{tabular}{@{}lcccccc@{}}
        \toprule
        {\bf Method}         & {\bf FID} $\downarrow$ & {\bf PR} $\uparrow$  & {\bf SSIM} $\uparrow$ & {\bf PSNR} $\uparrow$ & {\bf VIFp} $\uparrow$ & {\bf FSIM} $\uparrow$             \\ \midrule
        Immunization baseline (Random noise)  & 82.57 & 1.00 & $0.75 \pm 0.13$ & $19.21 \pm 4.00$ & $0.43 \pm 0.13$ & $0.83 \pm 0.08$  \\
        Immunization (Encoder attack)  & 130.6 & 0.95 & $0.58 \pm 0.11$ & $14.91 \pm 2.78$ & $0.30 \pm 0.10$ & $0.73 \pm 0.08$ \\
        Immunization (Diffusion attack) & \textbf{167.6} & \textbf{0.87} & $\mathbf{0.50 \pm 0.09} $  & $\mathbf{13.58 \pm 2.23}$ & $\mathbf{0.24 \pm 0.09}$ & $\mathbf{0.69 \pm 0.06}$ \\ \bottomrule
    \end{tabular}
    }
    \caption{We report various image quality metrics measuring the similarity between edits originating from immunized vs. non-immunized images. We observe that edits of immunized images are substantially different from those generated from the original (notn-immunized) images. Note that the arrows next to the metrics denote increasing image similarity. Since our goal is to make the edits as different as possible from the original edits in the presence of no immunization, then lower image similarity is better. Confidence intervals denote one standard deviation over 60 images. Additional metrics are in Appendix~\ref{app:additional_quantitative_results}.}
    \label{tab:similarity-metrics}
\end{table}

\subsection{Quantitative Results}
\paragraph{Image quality metrics.}
Figures~\ref{fig:image_var_simple_attack} and \ref{fig:image_edit_attacks} indicate that, as desired, edits of immunized images are noticeably different from those of non-immunized images. To quantify this difference, we generate 60 different edits of a variety of images using different prompts, and then compute several metrics capturing the similarity between resulting edits of immunized versus non-immunized images\footnote{We use the implementations provided in: \url{https://github.com/photosynthesis-team/piq}.}: FID~\citep{heusel2017gans}, PR~\citep{sajjadi18pr}, SSIM~\citep{wang04ssim}, PSNR, VIFp~\citep{sheikh06vifp}, and FSIM~\citep{zhang11fsim}\footnote{We report additional metrics in Appendix~\ref{app:additional_quantitative_results}.}. The better our immunization method is, the less similar the edits of immunized images are to those of non-immunized images. 

The similarity scores, shown in Table~\ref{tab:similarity-metrics}, indicate that applying  either of our immunization methods (encoder or diffusion attacks) indeed yields edits that are different from those of non-immunized images (since, for example, FID is far from zero for both of these methods).  As a baseline, we consider a naive immunization method that adds uniform random noise (of the same intensity as the perturbations used in our proposed immunization method). This method, as we verified, is not effective at disrupting the SDM, and yields edits almost identical to those of non immunized images. Indeed, in Table~\ref{tab:similarity-metrics}, the similarity scores of this baseline indicate closer edits to non-immunized images compared to both of our attacks.

\paragraph{Image-prompt similarity.} 

\begin{wrapfigure}{r}{0.5\textwidth}
    \centering
        \includegraphics[width=.4\columnwidth]{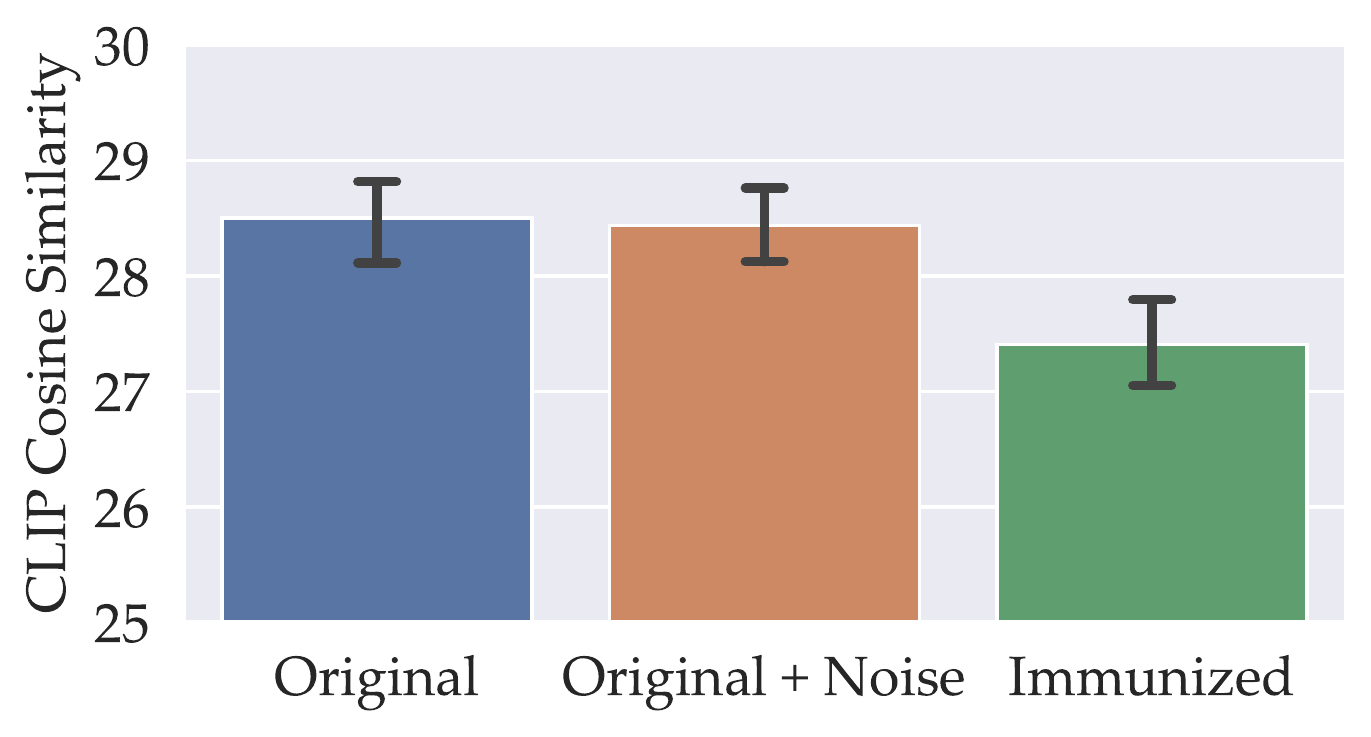}
        \caption{{\em Image-prompt similarity}. We plot the cosine similarity between the CLIP embeddings of the generated images and the text prompts, with and without immunization, as well as with a baseline immunization of adding small random noise to the original image. Error bars denote the interquartile range (IQR) over 60 runs.
        }
        \label{fig:barplot}
\end{wrapfigure}    
To further evaluate the quality of the generated/edited images after immunization (using diffusion attack), we measure the similarity between the edited images and the textual prompt used to guide this edit, with and without immunization. The fact that the SDM uses the textual prompt to guide the generation of an image indicates that the similarity between the generated image and the prompt should be high in the case of no immunization. However, after immunization (using the diffusion attack), the similarity should be low, since the immunization process disrupts the full diffusion process, and forces the diffusion model to ignore the prompt during generation. We use the same 60 edits as in our previous experiment, and we extract---using a pretrained CLIP model~\citep{radford2021learning}---the visual embeddings of these images and the textual prompts used to generate them. We then compute the cosine similarity between these two embeddings. As show in Figure~\ref{fig:barplot}, the immunization process decreases the similarity between the generated images and the textual prompts to generated them, as expected.

    \section{A Techno-Policy Approach to Mitigation of AI-Powered Editing}  
    \label{sec:discussion}
    In the previous sections we have developed an immunization procedure that, when applied to an image, protects the immunized version of that image from realistic manipulation by a given diffusion model.
Our immunization procedure has, however, certain important limitations.
We now discuss these limitations as well as a combination of technical and policy remedies needed to obtain a fully effective approach to raising the cost of malicious AI-powered image manipulation.

\paragraph{(Lack of) robustness to transformations.}
One of the limitations of our immunization method is that the adversarial perturbation that it relies on may be
ineffective after the immunized image is subjected to image transformations and noise purification techniques. For instance, malicious actors could attempt to remove the disruptive effect of that perturbation by cropping the image, adding filters to it, applying a rotation, or other means.
This problem can be addressed, however, by leveraging a long line of
research on creating {\em robust} adversarial perturbations, i.e., adversarial perturbations that can withstand a broad range of image modifications and noise manipulations~\cite{eykholt2018physical,kurakin2016adversarial,athalye2018synthesizing,brown2018adversarial}.

\paragraph{Forward-compatibility of the immunization.}
While the immunizing adversarial perturbations we produce might be effective at disrupting the current
generation of diffusion-based generative models, they are not guaranteed to be
effective against the future versions of these models. 
Indeed, one could hope to rely here on the so-called {\em transferability} of adversarial perturbations
\cite{papernot2016transferability,liu2017delving},
but {\em no} perturbation will be perfectly transferable.

To truly address this limitation, we thus need to go beyond purely technical methods and encourage---or compel---via policy means a collaboration between organizations that develop large diffusion models, end-users, as well as data hosting and dissemination platforms. Specifically, this collaboration would involve the developers providing APIs that allow the users and platforms to immunize their images against manipulation by the diffusion models the developers create. Importantly, these APIs should guarantee ``forward compatibility'', i.e., effectiveness of the offered immunization against models developed in the future. This can be accomplished by planting, when training such future models, the current immunizing adversarial perturbations as backdoors. (Observe that our immunization approach can provide {\em post-hoc} ``backward compatibility'' too. That is, one can create immunizing adversarial perturbations that are effective for models that were {\em already released}.)

It is important to point out that we are leveraging here an incentive alignment that is fundamentally different to the one present in more typical applications of adversarial perturbations and backdoor attacks. In particular, the ``attackers'' here---that is, the parties that create the adversarial perturbations/execute the backdoor attack---are the same parties that develop the models being attacked. This crucial difference is, in particular, exactly what helps remedy the forward compatibility challenges that turns out to be crippling, e.g., in the context of ``unlearnable'' images creation (i.e., creation of images that are immune to being leveraged by, e.g., facial recognition models) \cite{dixit21poisonunlearn}.

    \section{Related Work}
    \label{sec:related}
    
\paragraph{Data misuse after model training.} Recent advances in ML-powered
image generation and editing have raised concerns about the potential misuse of
personal data for generating fake images. This issue arose first in the context of the
development of generative adversarial networks (GANs) for image generation and
editing~\citep{goodfellow2014generative,mirza2014conditional, salimans2016improved, isola2017image, zhu2017unpaired, zhang2017stackgan, karras2018progressive, brock2019large, karras2019style}, and led to research on
methods for defending against such manipulation, such as attacking the GAN
itself~\citep{ruiz2020disrupting, ruiz20protecting, sun2021adversarial}.
This problem became exacerbated recently with the advent of (publicly available) diffusion models~\cite{rombach2022high, ramesh2022hierarchical}. Indeed, one can now easily describe in text how one wants to manipulate an image, and immediately get the result of an impressive quality (see Figure \ref{fig:applications_of_diff_models}) that significantly outperforms previous methods, such as GANs.

\paragraph{Deepfake detection.} A line of work related to ours aims to
{\em detect} fake images rather than {\em prevent} their generation. Deepfake
detection methods include analyzing the consistency of facial expressions and
identifying patterns or artifacts in the image that may indicate manipulation, and training machine learning models to recognize fake images~\cite{korshunov2018deepfakes, afchar2018mesonet, nguyen19deepfakesurvey, mirsky2021creation, rossler2019faceforensics++, durall2019unmasking, li2020face, li2020celeb, bonettini2021video}. While some deepfake detection methods are more effective than others, no single method is foolproof.
A potential way to mitigate this shortcoming could involve development of so-called watermarking methods~\cite{cox1997secure, neekhara2022facesigns}. These methods aim to ensure that it is easy to detect that a given output has been produced using a generative model---such watermarking approaches have been recently developed for a related context of large language models \cite{kirchenbauer23lmwatermark}. 
Still, neither deepfake detection nor watermarking methods could
protect images from being manipulated in the first place. 
A manipulated image can hence cause harm before being flagged as fake. Also, given that our work is complementary to deepfake detection and watermarking methods, it could, in principle, be combined with them.

\paragraph{Data misuse during model training.} The abundance of readily available data on the Internet has played a significant role in recent breakthroughs in deep learning, but has also raised concerns about the potential misuse of such data when training models. Therefore, there has been an increasing interest in protection against unauthorized data
exploitation, e.g., by designing unlearnable
examples~\citep{huang2021unlearnable, fu2021robust}. These methods propose
adding imperceptible backdoor signals to user data before uploading it online, so as to prevent models from fruitfully utilizing this data. However, as pointed out by~\citet{dixit21poisonunlearn}, these methods can be circumvented, often simply by waiting until subsequently developed models can avoid being fooled by the planted backdoor signal.

    \section{Conclusion}
    \label{sec:conclusion}
    In this paper, we presented a method for raising the difficulty of using diffusion models for malicious image manipulation. Our method involves ``immunizing'' images through addition imperceptible adversarial perturbations. These added perturbations disrupt the inner workings of the targeted diffusion model and thus prevent it from producing realistic modifications of the immunized images.

We also discussed the complementary policy component that will be needed to make our approach fully effective. This component involves ensuring cooperation of the organizations developing diffusion-based generative models in provisioning APIs that allow users to immunize their images to manipulation by such models (and the future versions of thereof).

    \section{Acknowledgements}
    Work supported in part by the NSF grants CNS-1815221 and DMS-2134108, and Open Philanthropy. This material is based upon work supported by the Defense Advanced Research Projects Agency (DARPA) under Contract No. HR001120C0015. Work partially done on the MIT Supercloud compute cluster~\citep{reuther2018interactive}.

    \clearpage
    \printbibliography
    \clearpage

    \appendix
    \onecolumn
    \section{Experimental Setup}
    \label{app:experimental}

\subsection{Details of the diffusion model we used}
In this paper, we used the open-source stable diffusion model hosted on the Hugging Face\footnote{This model is available on: \url{https://huggingface.co/runwayml/stable-diffusion-v1-5}.}. We use the hyperparameters presented in Table~\ref{tab:diff_hparams} to generated images from this model. For a given image on which we want to test our immunization method, we first search for a good random seed that leads to a realistic modification of the image given some textual prompt. Then we use the same seed when editing the immunized version of the same image using the diffusion model. This ensures that the immunized image is modified in the same way as the original image, and that the resulting non-realistic edits are due to immunization and not to random seed.

\begin{table}[!h]
    \centering
    \caption{Hyperparameters used for the Stable Diffusion model.}
    \label{tab:diff_hparams}
    \begin{tabular}{@{}ccccc@{}}
    \toprule
    {\bf height} & {\bf width} & {\bf guidance\_scale} & {\bf num\_inference\_steps} & {\bf eta} \\ \midrule
    512    &  512  & 7.5             & 100                   & 1   \\ \bottomrule
    \end{tabular}
\end{table}

\subsection{Our attacks details}
Throughout the paper, we use two different attacks: an encoder attack and a diffusion attack. These attacks are described in the main paper, and are summarized here in Algorithm~\ref{alg:simple_attack_alg} and Algorithm~\ref{alg:complex_attack_alg}, respectively. For both of the attacks, we use the same set of hyperparameters shown in Table~\ref{tab:attack_hparams}.
The choice of $\epsilon$ was such that it is the large enough to disturb the image, but small enough to not be noticeable by the human eye.

\begin{table}[!h]
    \centering
    \caption{Hyperparameters used for the adversarial attacks.}
    \label{tab:attack_hparams}
    \begin{tabular}{@{}cccc@{}}
    \toprule
    {\bf Norm}          & $\bm{\epsilon}$ & {\bf step size} & {\bf number of steps} \\ \midrule
    $\ell_\infty$ & 16/255 & 2/255      & 200             \\ \bottomrule
    \end{tabular}
\end{table}

\begin{algorithm}[!hbtp]
    \caption{Encoder Attack on a Stable Diffusion Model}
    \label{alg:simple_attack_alg}
    \begin{algorithmic}[1]
        \STATE {\bfseries Input:} Input image $\xd$, target image $\xd_{targ}$, Stable Diffusion model encoder $\mathcal{E}$, perturbation budget $\epsilon$, step size $k$, number of steps $N$.
        \STATE Compute the embedding of the target image: $\zd_{targ} \leftarrow \mathcal{E}(\xd_{targ})$
        \STATE Initialize adversarial perturbation $\delta_{encoder} \leftarrow 0$, and immunized image $\xd_{im} \leftarrow \xd$
        \FOR{$n = 1 \ldots N$}
            \STATE Compute the embedding of the immunized image: $\zd \leftarrow \mathcal{E}(\xd_{im})$
            \STATE Compute mean squared error: $l \leftarrow \lVert \zd_{targ} - \zd \rVert_2^2$
            \STATE Update adversarial perturbation:
            $\delta_{encoder} \leftarrow \delta_{encoder} + k\cdot \text{sign}(\nabla_{\xd_{im}} l)$
            \STATE $\delta_{encoder} \leftarrow \text{clip}(\delta_{encoder}, -\epsilon, \epsilon)$
            \STATE Update the immunized image: $\xd_{im} \leftarrow \xd_{im} - \delta_{encoder}$
        \ENDFOR
        \STATE {\bfseries Return:} $\xd_{im}$
    \end{algorithmic}
\end{algorithm}

\begin{algorithm}[!hbtp]
    \caption{Diffusion Attack on a Stable Diffusion Model}
    \label{alg:complex_attack_alg}
    \begin{algorithmic}[1]
        \STATE {\bfseries Input:} Input image $\xd$, target image $\xd_{targ}$, Stable Diffusion model $f$, perturbation budget $\epsilon$, step size $k$, number of steps $N$.
        \STATE Initialize adversarial perturbation $\delta_{diffusion} \leftarrow 0$, and immunized image $\xd_{im} \leftarrow \xd$
        \FOR{$n = 1 \ldots N$}
            \STATE Generate an image using diffusion model: $\xd_{out} \leftarrow f(\xd_{im})$
            \STATE Compute mean squared error: $l \leftarrow \lVert \xd_{targ} - \xd_{out} \rVert_2^2$
            \STATE Update adversarial perturbation:
            $\delta_{diffusion} \leftarrow \delta_{diffusion} + k\cdot\text{sign}(\nabla_{\xd_{im}} l)$
            \STATE $\delta_{diffusion} \leftarrow \text{clip}(\delta_{diffusion}, -\epsilon, \epsilon)$
            \STATE Update the immunized image: $\xd_{im} \leftarrow \xd_{im} - \delta_{diffusion}$
        \ENDFOR
        \STATE {\bfseries Return:} $\xd_{im}$
    \end{algorithmic}
\end{algorithm}

    \section{Extended Background for Diffusion Models}
    \label{app:extended_background}

\paragraph{Overview of the diffusion process.}
At their heart, diffusion models leverage a statistical concept: the diffusion process~\citep{sohl-dickstein15diffusion, ho2020ddpm}. Given a sample $\xd_0$ from a distribution of real images $q(\cdot)$, the diffusion process works in two steps: a forward step and a backward step. During the forward step, Gaussian noise is added to the sample $\xd_0$ over $T$ time steps, to generate increasingly noisier versions $\xd_1, \ldots, \xd_{T}$ of the original sample $\xd_0$, until the sample is equivalent to an isotropic Gaussian distribution. During the backward step, the goal is to reconstruct the original sample $\xd_0$ by iteratively denoising the noised samples $\xd_T, \ldots, \xd_1$. The power of the diffusion models stems from the ability to learn the backward process using neural networks. This allows to generate new samples from the distribution $q(\cdot)$ by first generating a random Gaussian sample, and then passing it through the ``neural'' backward step.

\paragraph{Forward process.}
\vskip -0.2cm
During the forward step, Gaussian noise is iteratively added to the original sample $\xd_0$. The forward process $q(\xd_{1:T} | \xd_0)$ is assumed to follow a Markov chain, i.e. the sample at time step $t$ depends only on the sample at the previous time step. Furthermore, the variance added at a time step $t$ is controlled by a schedule of variances $\{\beta_t \}_{t=1}^T$\footnote{The values of $a_t$ and $b_t$ from the main paper correspond to $a_t = \sqrt{1-\beta_t}$ and $b_t = \beta_t$}.

\begin{equation}
    \label{eq:forward_process}
    \begin{aligned}
        q(\xd_{1:T} | \xd_0) = \prod_{t=1}^T q(\xd_t | \xd_{t-1}); & \qquad q(\xd_t | \xd_{t-1}) = \mathcal{N}(\xd_t; \sqrt{1-\beta_t} \xd_{t-1}, \beta_t \mathbf{I})
    \end{aligned}
\end{equation}

\paragraph{Backward process.}
\vskip -0.2cm
At the end of the forward step, the sample $\xd_T$ looks as if it is sampled from an isotropic Gaussian $p(\xd_T) = \mathcal{N}(\xd_T; \mathbf{0}, \mathbf{I})$. Starting from this sample, the goal is to recover $\xd_{0}$ by iteratively removing the noise using neural networks. The joint distribution $p_\theta(\xd_{0:T})$ is referred to as the reverse process, and is also assumed to be a Markov chain.

\begin{equation}
    \label{eq:backward_process}
    \begin{aligned}
        p_\theta(\xd_{0:T}) = p(\xd_T) \prod_{t=1}^T p_\theta(\xd_{t-1} | \xd_t); & \qquad p_\theta(\xd_{t-1} | \xd_t) = \mathcal{N}(\xd_{t-1}; \bm{\mu}_\theta(\xd_t, t), \bm{\Sigma}_\theta (\xd_t, t))
    \end{aligned}
\end{equation}

\paragraph{Training a diffusion model.}
\vskip -0.2cm
At its heart, diffusion models are trained in a way similar to Variational Autoencoders, i.e. by optimizing a variational lower bound. Additional tricks are employed to make the process faster. For an extensive derivation, refer to~\citep{weng2021diffusion}.

\begin{equation}
    \label{eq:vlb}
    \begin{aligned}
        \mathbb{E}_{q(\xd_0)}[- \log p_\theta (\xd_0)] \leq \mathbb{E}_{q(\xd_{0:T})} \left[ \log \frac{q(\xd_{1:T} | \xd_0)}{p_\theta (\xd_{0:T})} \right] = L_{VLB}
    \end{aligned}
\end{equation}

\paragraph{Latent Diffusion Models (LDMs).}
\vskip -0.2cm
 In this paper, we focus on a specific class of diffusion models, namely LDMs, which was proposed in~\citep{rombach2022high} as a model that applies the diffusion process described above in a \textit{latent space} instead of the image space. This enables efficient training and inference of diffusion models.

To train an LDM, the input image $\xd_0$ is first mapped to a latent representation $\zd_0 = \mathcal{E}(\xd_0)$, where $\mathcal{E}$ is an image encoder. This input representation $\zd_0$ is then passed to the diffusion process to obtain a denoised $\tilde{\zd}$. The generated image $\tilde{\xd}$ is then obtained by decoding $\tilde{\zd}_0$ using a decoder $\mathcal{D}$, i.e. $\tilde{\xd} = \mathcal{D}(\tilde{\zd})$.

    \section{Additional Results}
    \label{app:additional_results}
    \subsection{Additional quantitative results}
\label{app:additional_quantitative_results}

We presented in Section~\ref{sec:results} several metrics to assess the similarity between the images generated with and without immunization. Here, we report in Table~\ref{tab:additional-similarity-metrics} additional metrics to evaluate this: SR-SIM~\citep{zhang12srsim}, GMSD~\citep{xue14gmsd}, VSI~\citep{zhang14vsi}, DSS~\citep{balanov15dss}, and HaarPSI~\citep{reisenhofer18haarpsi}. Similarly, we indicate for each metric whether a higher value corresponds to higher similarity (using $\uparrow$), or contrariwise (using $\downarrow$). We again observe that applying the encoder attack already decreases the similarity between the generated images with and without immunization, and applying the diffusion attack further decreases the similarity.

\begin{table}[!h]
    \centering
    \caption{Additional similarity metrics for Table~\ref{tab:similarity-metrics}. Errors denote standard deviation over 60 images.}
    \label{tab:additional-similarity-metrics}
    \begin{tabular}{@{}lccccc@{}}
        \toprule
        {\bf Method}         & {\bf SR-SIM} $\uparrow$ & {\bf GMSD} $\downarrow$ & {\bf VSI} $\uparrow$ & {\bf DSS} $\uparrow$ & {\bf HaarPSI} $\uparrow$             \\ \midrule
        Immunization baseline (Random noise)  & $0.91 \pm 0.04$ & $0.20 \pm 0.06$ & $0.94 \pm 0.03$ & $0.35 \pm 0.18$ & $0.52 \pm 0.15$  \\
        Immunization (Encoder attack)  & $0.86 \pm 0.05$ & $0.26 \pm 0.05$ & $0.90 \pm 0.03$ & $0.19 \pm 0.09$ & $0.35 \pm 0.11$ \\
        Immunization (Diffusion attack) & $\mathbf{0.84 \pm 0.05} $  & $\mathbf{0.27 \pm 0.04}$ & $\mathbf{0.89 \pm 0.03}$ & $\mathbf{0.17 \pm 0.08}$ & $\mathbf{0.31 \pm 0.08}$ \\ \bottomrule
    \end{tabular}
\end{table}

\clearpage

\subsection{Generating Image Variations using Textual Prompts}
\begin{figure}[!h]
    \centering
    \includegraphics[width=0.9\columnwidth]{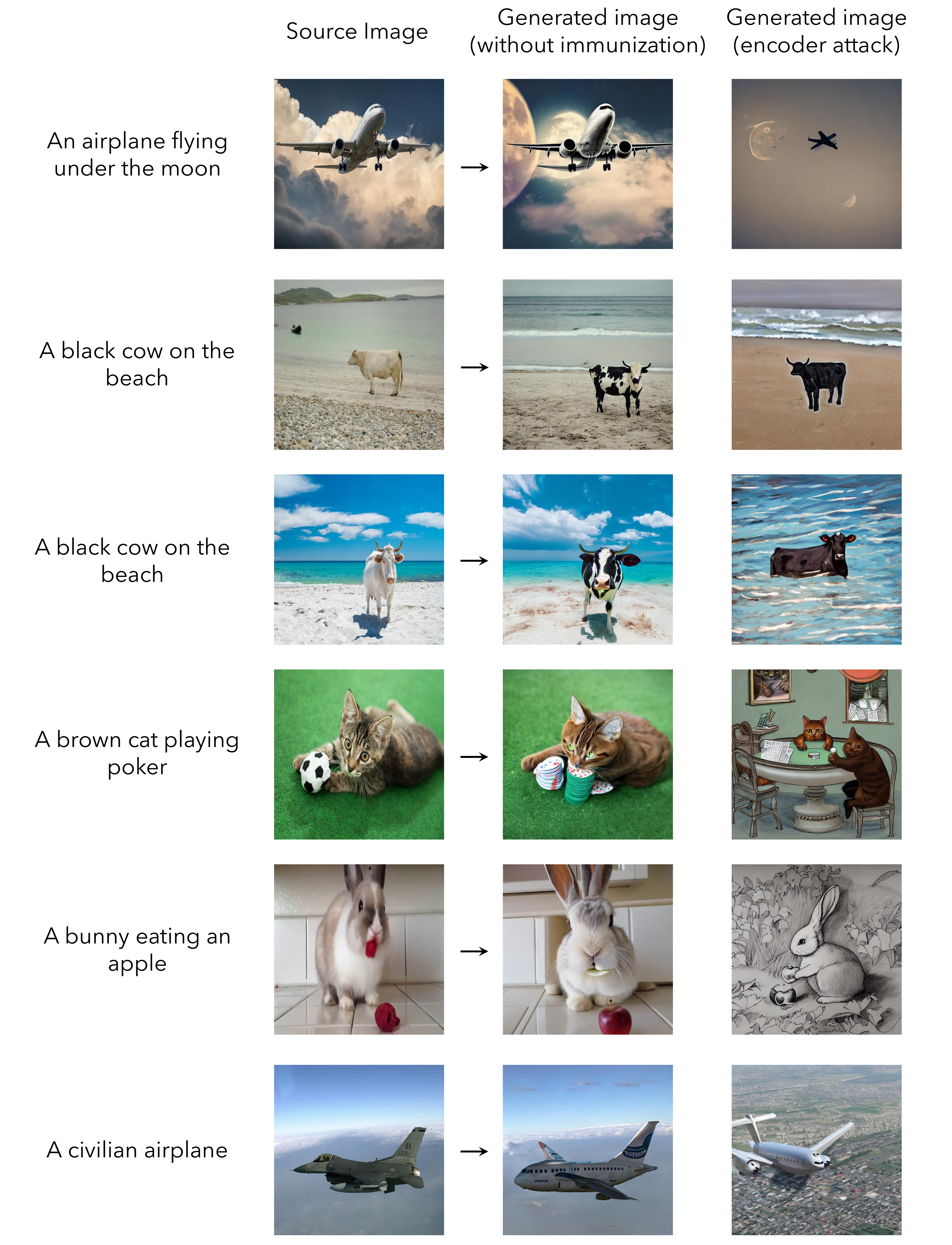}
    \caption{Immunization against generating prompt-guided image variations.}
    \label{fig:app_image_var}
\end{figure}

\newpage

\subsection{Image Editing via Inpainting}

\begin{figure}[!h]
    \centering
    \includegraphics[width=\columnwidth]{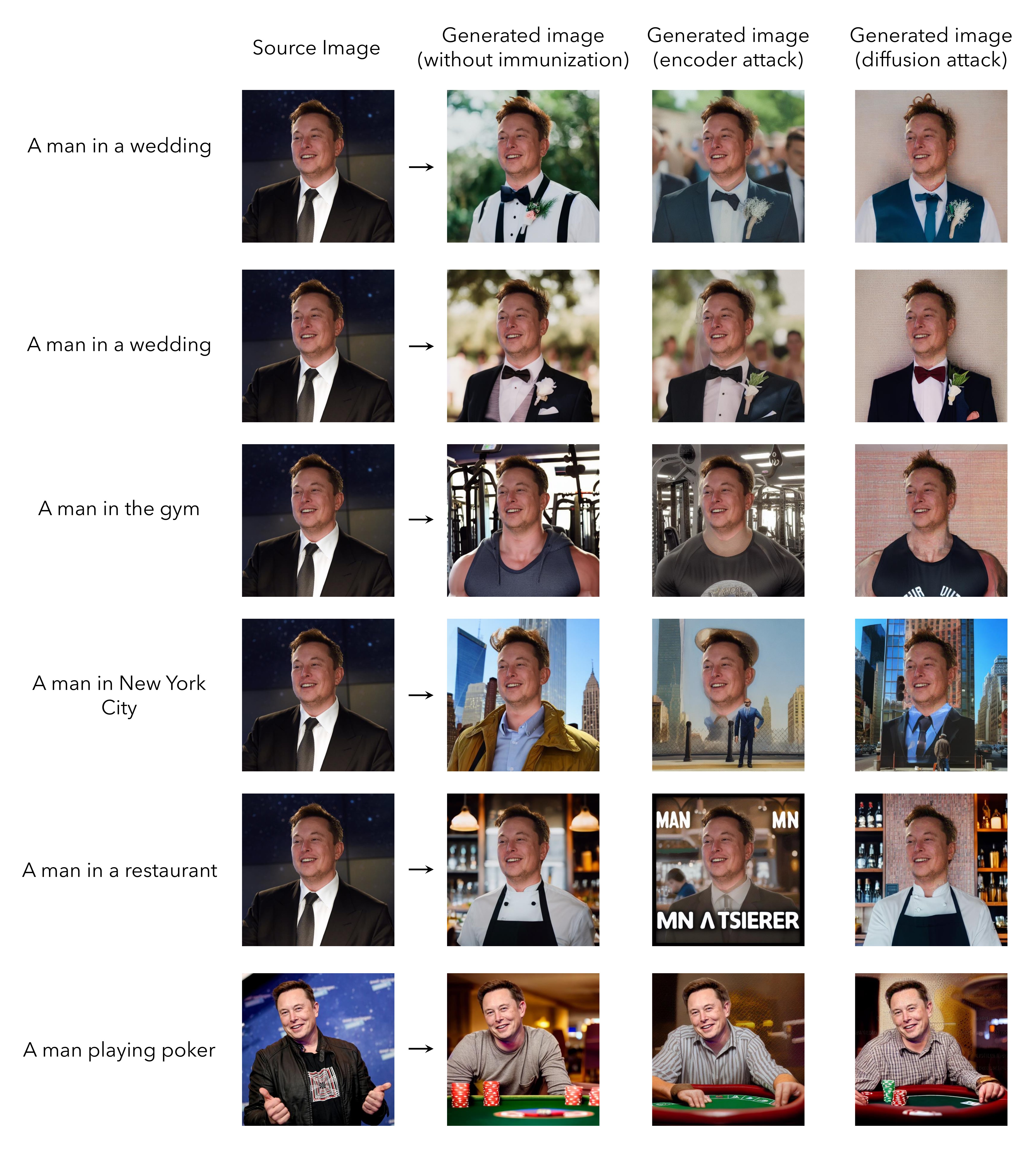}
    \caption{Immunization against image editing via prompt-guided inpainting.}
    \label{fig:app_image_editing_1}
\end{figure}

\begin{figure}[!h]
    \centering
    \includegraphics[width=\columnwidth]{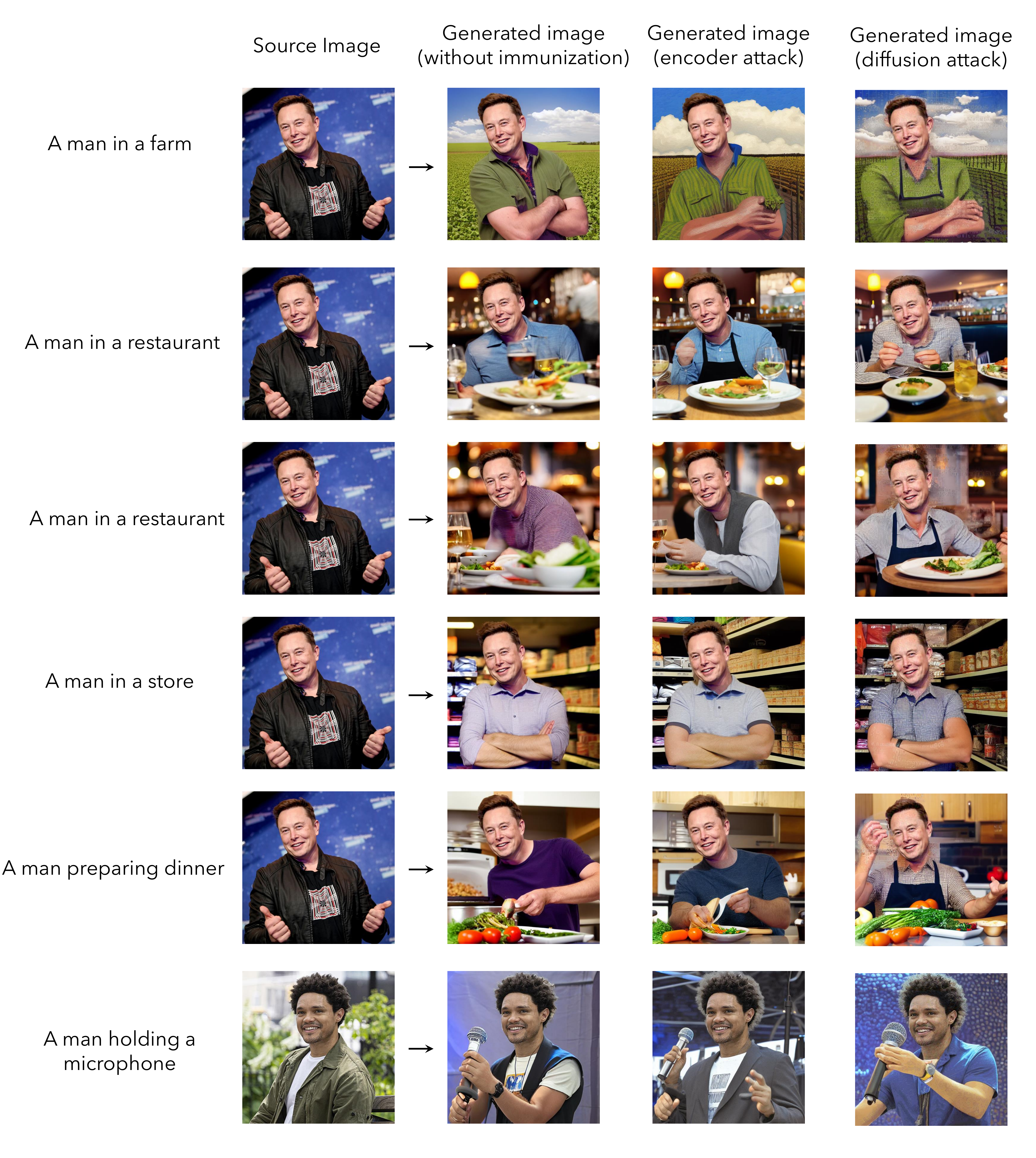}
    \caption{Immunization against image editing via prompt-guided inpainting.}
    \label{fig:app_image_editing_2}
\end{figure}

\begin{figure}[!h]
    \centering
    \includegraphics[width=\columnwidth]{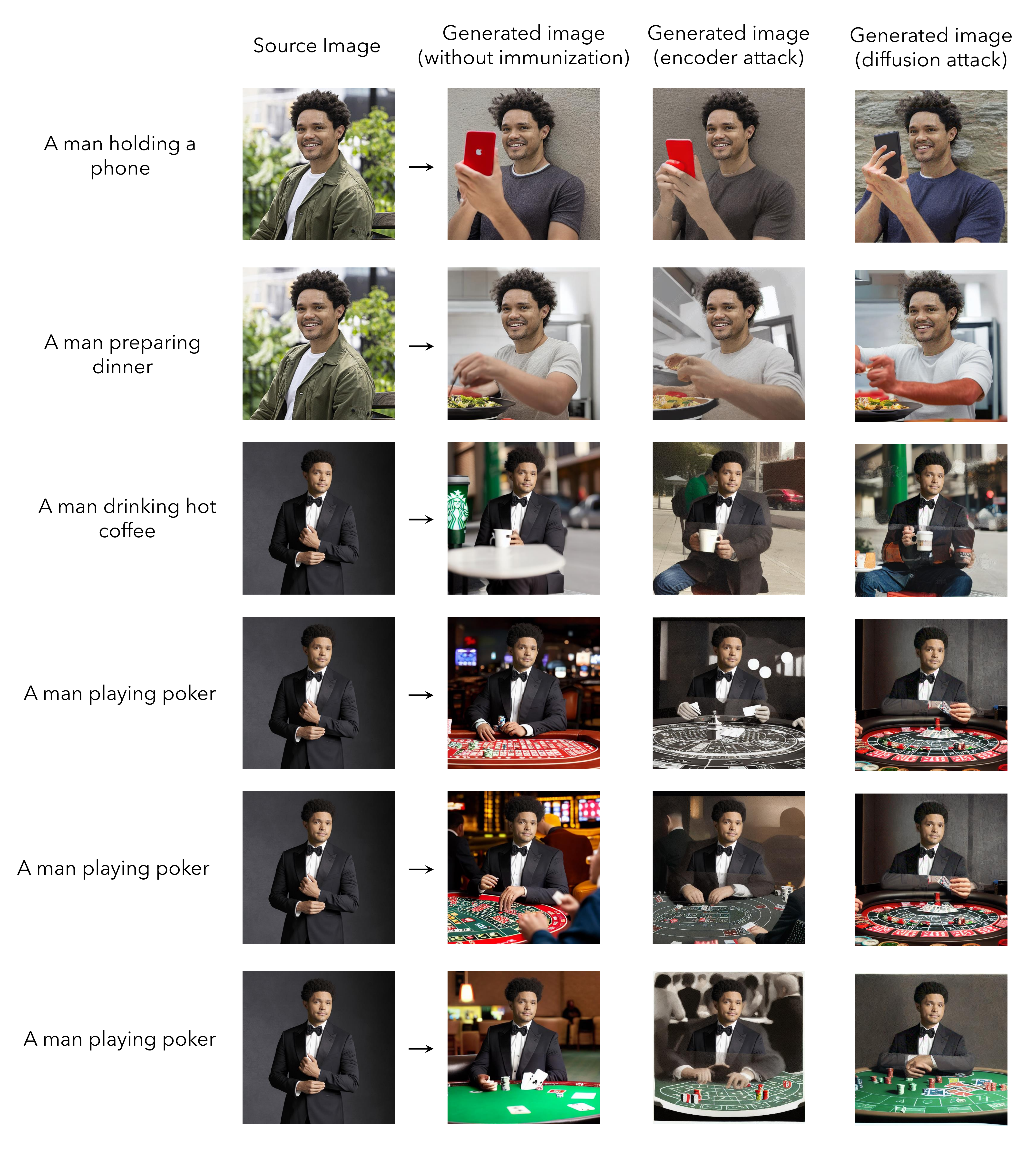}
    \caption{Immunization against image editing via prompt-guided inpainting.}
    \label{fig:app_image_editing_3}
\end{figure}

\begin{figure}[!h]
    \centering
    \includegraphics[width=\columnwidth]{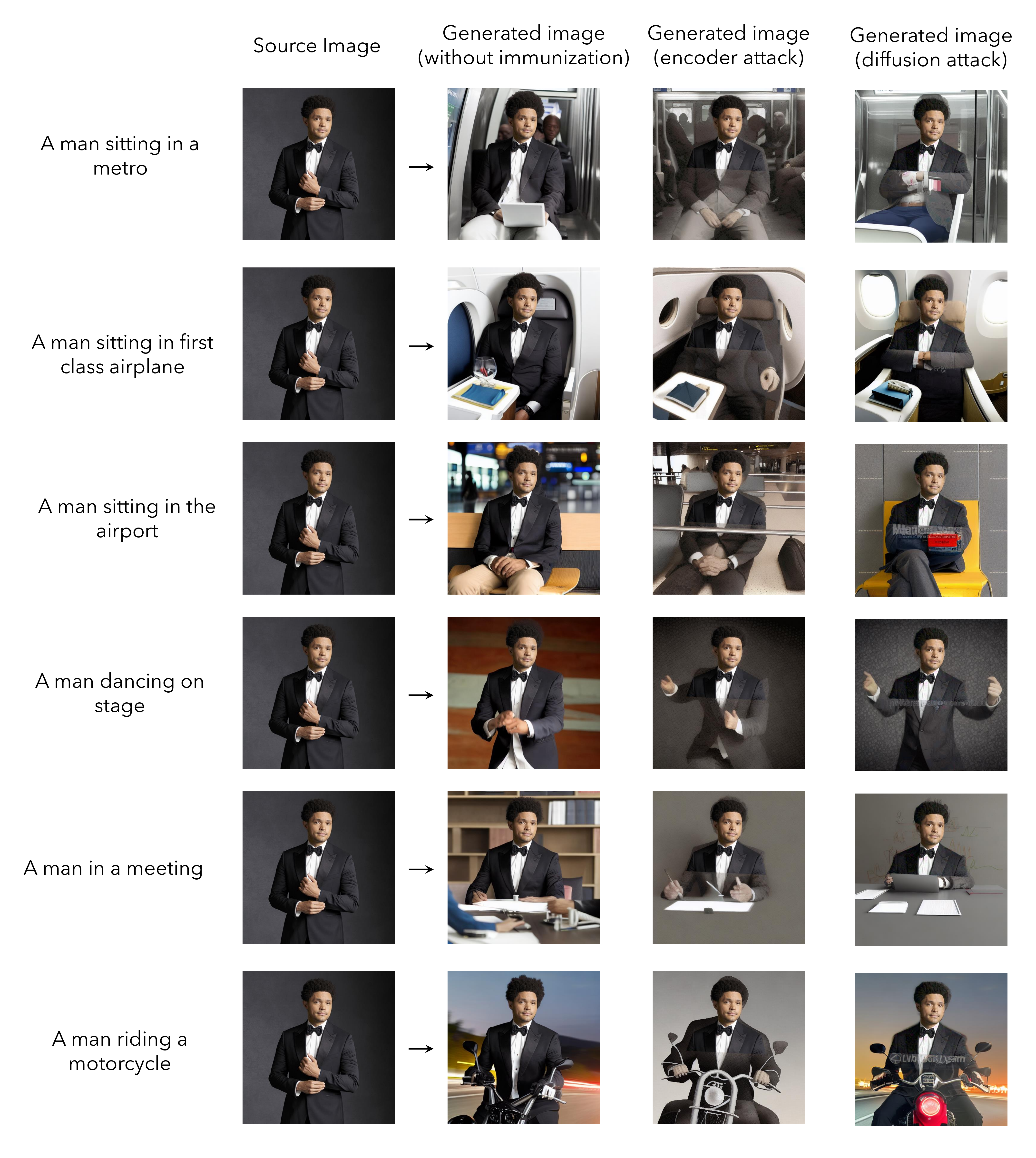}
    \caption{Immunization against image editing via prompt-guided inpainting.}
    \label{fig:app_image_editing_4}
\end{figure}

\begin{figure}[!h]
    \centering
    \includegraphics[width=\columnwidth]{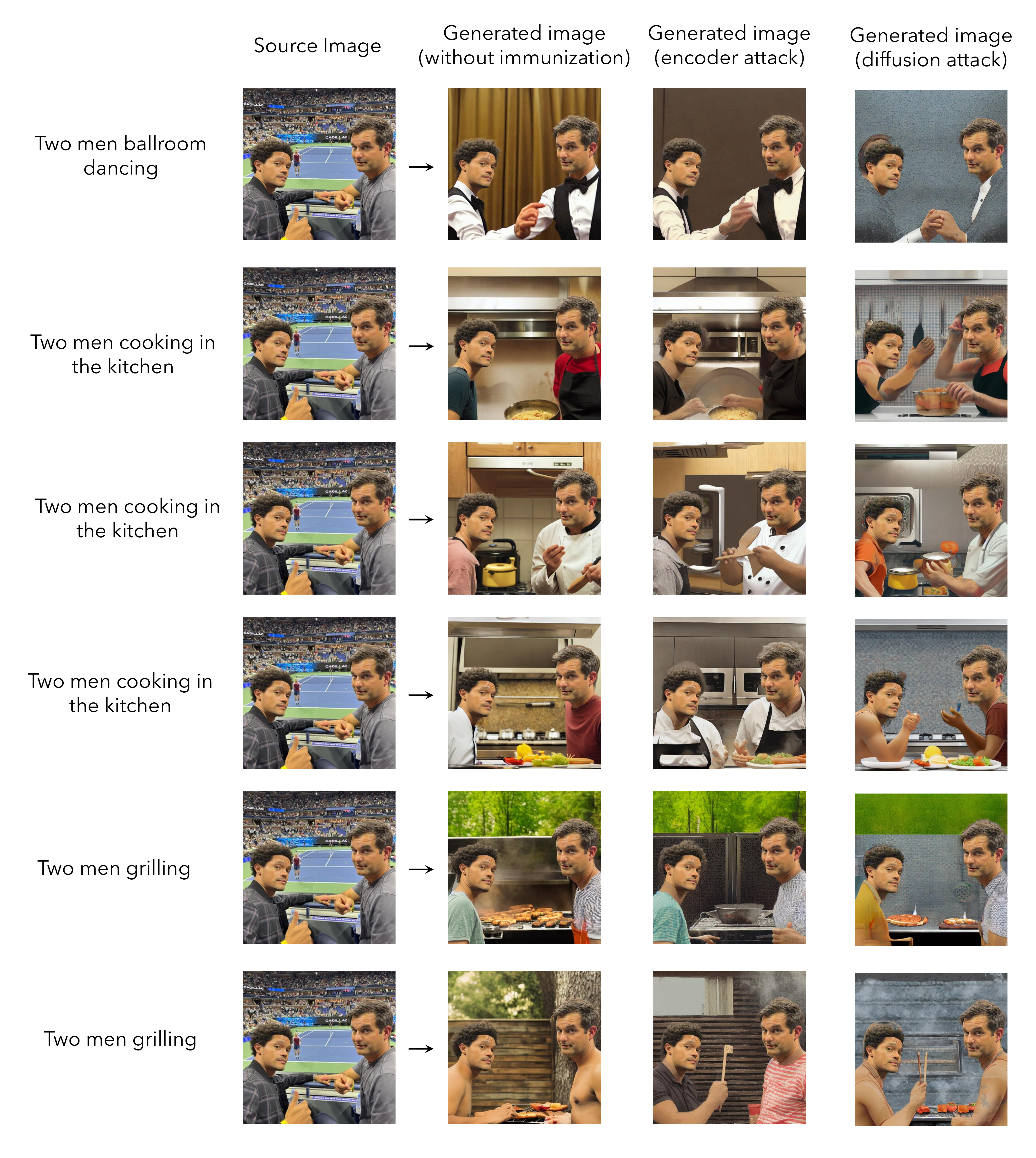}
    \caption{Immunization against image editing via prompt-guided inpainting.}
    \label{fig:app_image_editing_5}
\end{figure}

\begin{figure}[!h]
    \centering
    \includegraphics[width=\columnwidth]{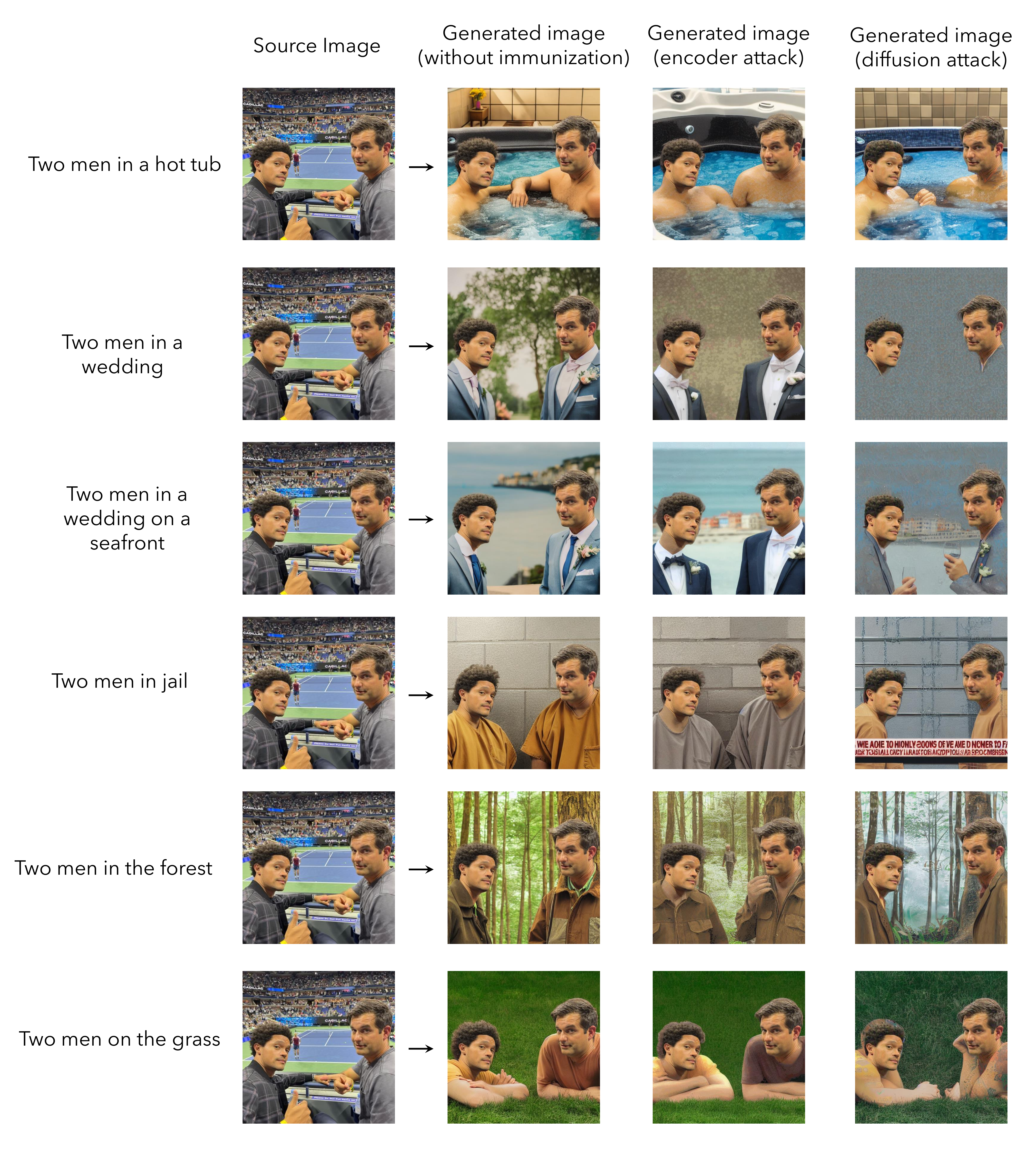}
    \caption{Immunization against image editing via prompt-guided inpainting.}
    \label{fig:app_image_editing_6}
\end{figure}

\begin{figure}[!h]
    \centering
    \includegraphics[width=\columnwidth]{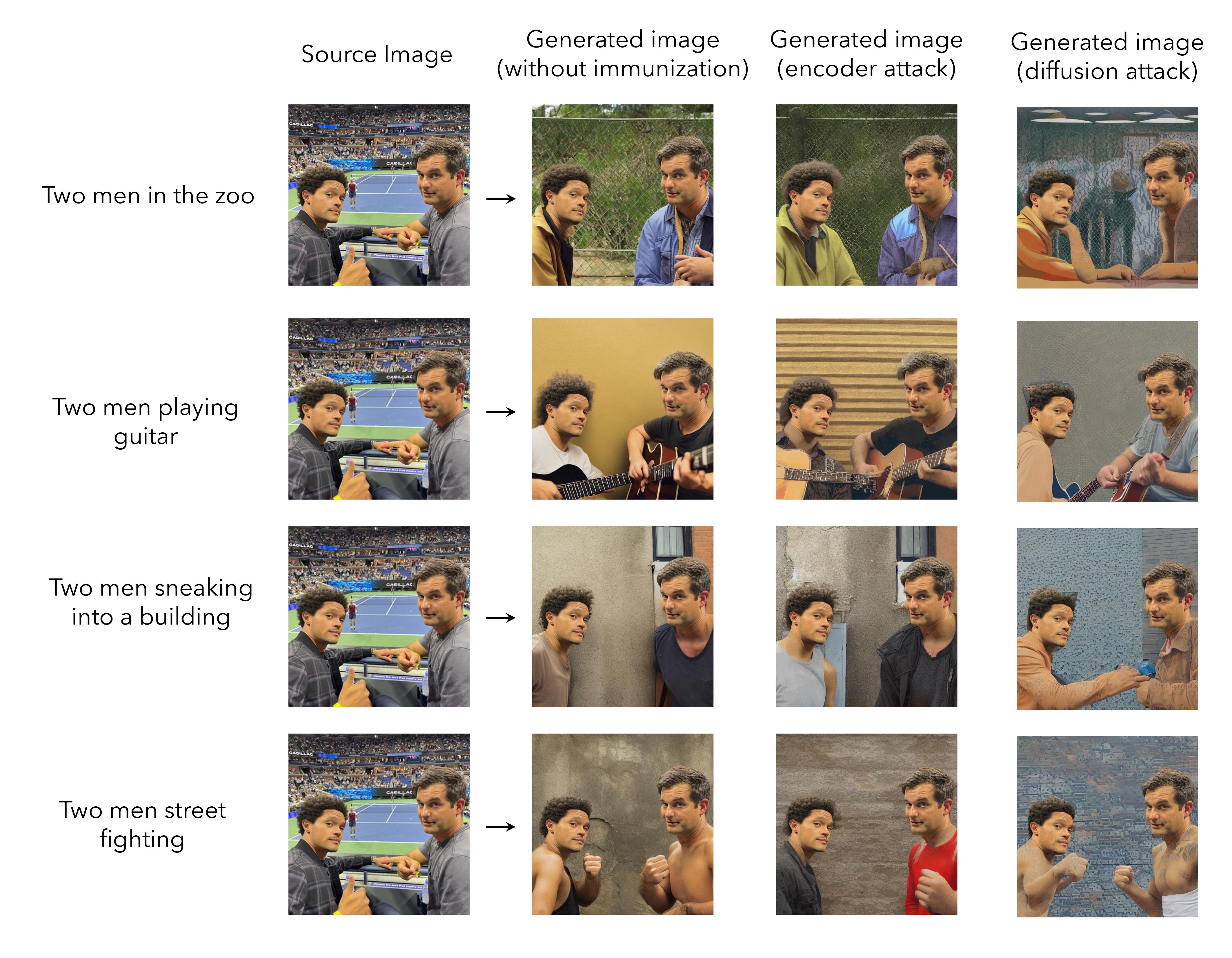}
    \caption{Immunization against image editing via prompt-guided inpainting.}
    \label{fig:app_image_editing_7}
\end{figure}

\begin{figure}[!h]
    \centering
    \includegraphics[width=\columnwidth]{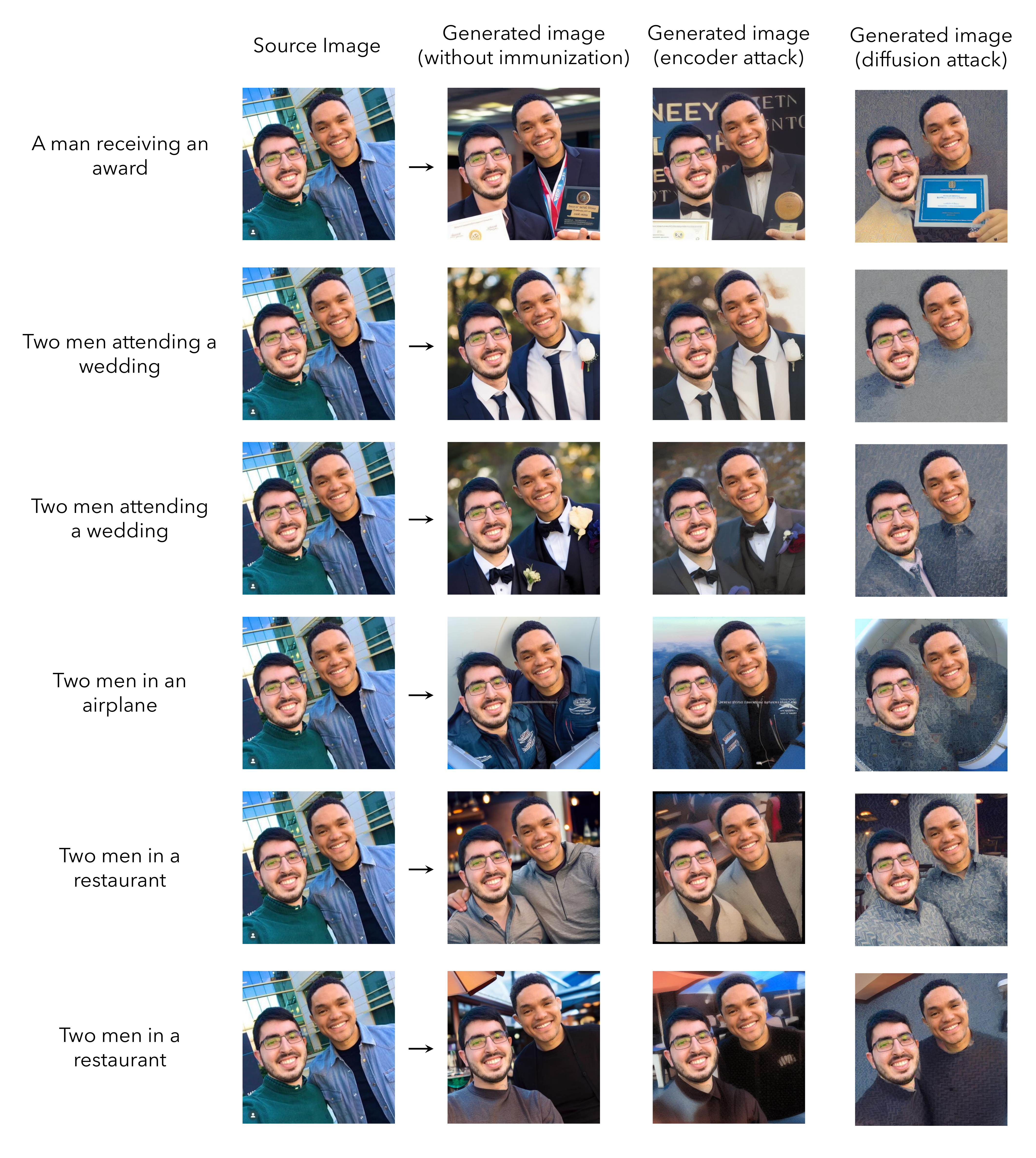}
    \caption{Immunization against image editing via prompt-guided inpainting.}
    \label{fig:app_image_editing_8}
\end{figure}

\begin{figure}[!h]
    \centering
    \includegraphics[width=\columnwidth]{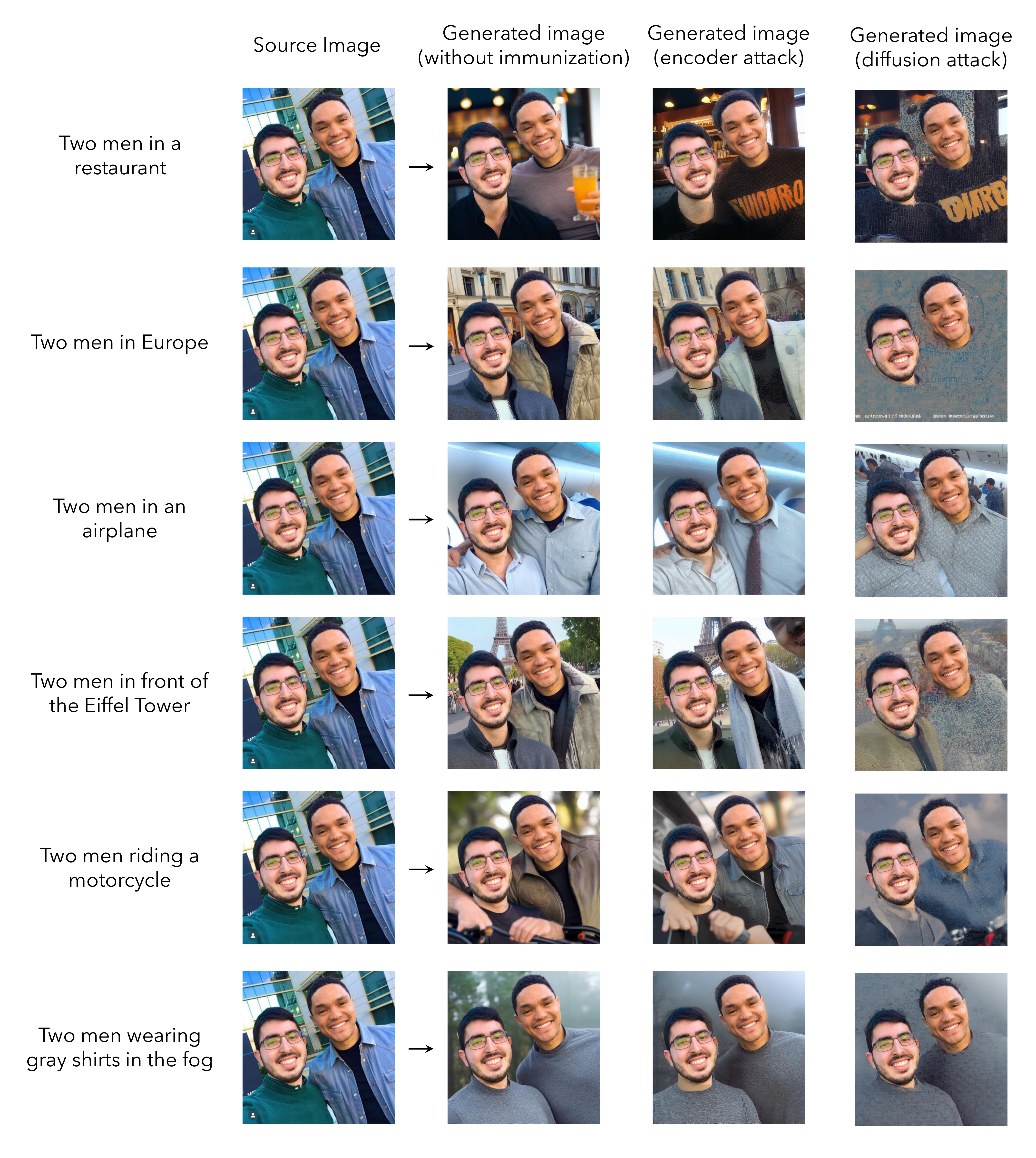}
    \caption{Immunization against image editing via prompt-guided inpainting.}
    \label{fig:app_image_editing_9}
\end{figure}

\begin{figure}[!h]
    \centering
    \includegraphics[width=\columnwidth]{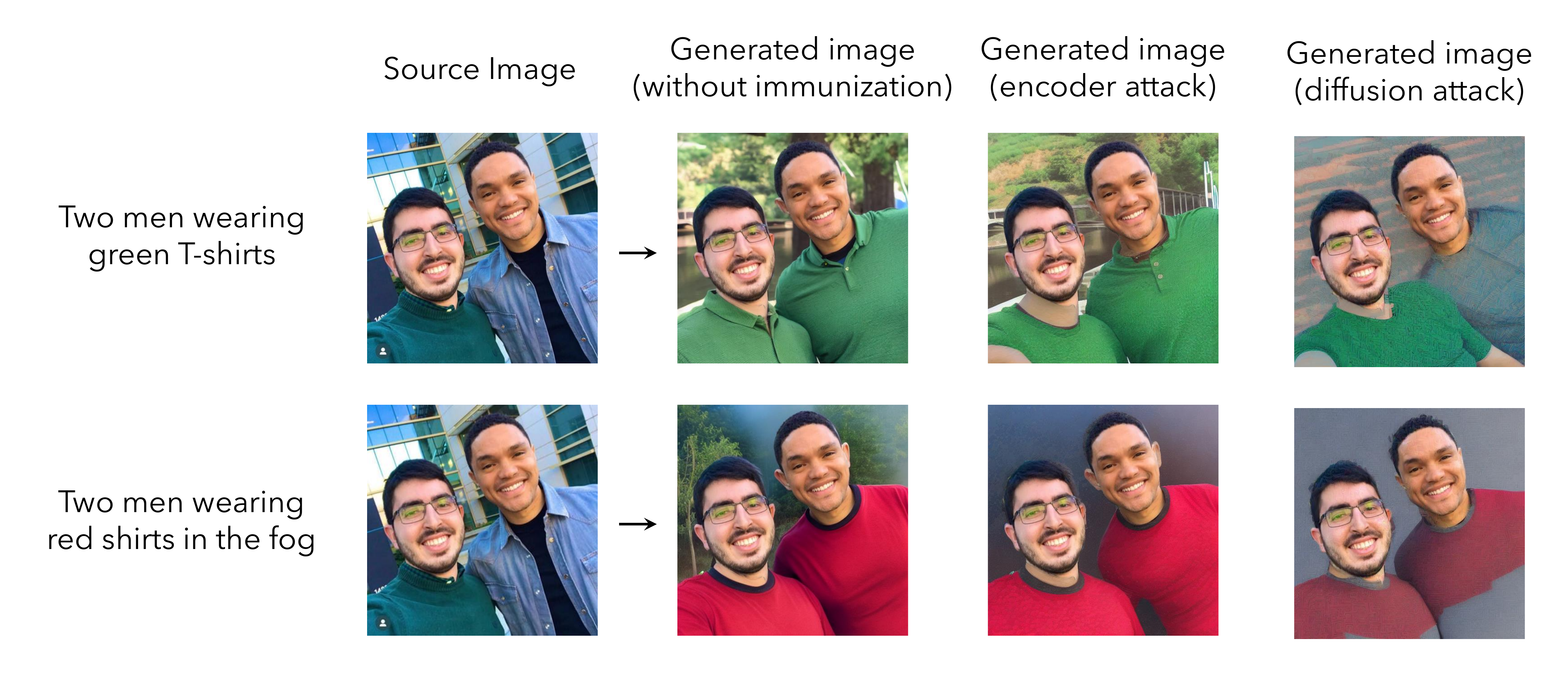}
    \caption{Immunization against image editing via prompt-guided inpainting.}
    \label{fig:app_image_editing_10}
\end{figure}

    \end{document}